\pdfoutput=1

\documentclass[11pt]{article}

\usepackage[preprint]{acl}

\usepackage{times}
\usepackage{lipsum}
\usepackage{natbib}
\usepackage{amssymb}
\usepackage{latexsym}
\usepackage{float}
\usepackage{subcaption}
\usepackage{graphicx}
\usepackage{hyperref}
\usepackage{multirow}
\usepackage{tabularx}
\usepackage{enumitem}
\usepackage{array}
\usepackage{amsmath}
\usepackage[export]{adjustbox}

\usepackage[T1]{fontenc}

\usepackage[utf8]{inputenc}

\usepackage{microtype}
\usepackage{graphicx}
\usepackage{listings}

\usepackage{inconsolata}

\usepackage[ruled,longend]{algorithm2e}
\usepackage{amssymb}

\usepackage{xcolor}

\usepackage{tcolorbox}
\usepackage{tikz}
\tcbuselibrary{listings}
\definecolor{lightgray}{HTML}{F2F2F2}
\definecolor{darkgray}{HTML}{333333}

\newcommand{\roundedbox}{%
    \tikz{\draw[line width=0.4pt,rounded corners=1.5pt] (0,0) rectangle (0.21,0.21);}%
}


\usepackage{tcolorbox}
\newcounter{promptno}[section]
\newlength\mystoreparindent
\newenvironment{prompt}[1][]
{
  \setlength{\mystoreparindent}{\the\parindent}
  \setlength{\parindent}{0pt}
  \refstepcounter{promptno}
  \par\medskip
  \noindent
  \begin{tcolorbox}[left=1pt,right=1pt]
  \small
  \tt
}{
  \end{tcolorbox}
  \setlength{\parindent}{\mystoreparindent}
  \medskip
}

\usepackage{color}

\usepackage{booktabs}
\newcolumntype{T}{>{\ttfamily}l}

\newcommand{\model}[1]{\texttt{#1}}

\title{\textit{How Many Parameters Does it Take to Change a Light Bulb?}\\
Evaluating Performance in Self-Play of Conversational \\ Games as a Function of Model Characteristics}

\author{
 \textbf{Nidhir Bhavsar\textsuperscript{1}},
 \textbf{Jonathan Jordan\textsuperscript{1}},
 \textbf{Sherzod Hakimov\textsuperscript{1}},
 \textbf{David Schlangen\textsuperscript{1,2}}
\\
\\
 \textsuperscript{1}Computational Linguistics, Department of Linguistics\\University of Potsdam, Germany\\
 \textsuperscript{2}German Research Center for Artificial Intelligence (DFKI), Berlin, Germany
\\
\texttt{firstname.lastname@uni-potsdam.de} 
}

\begin{document}
\maketitle
\begin{abstract}
What makes a good Large Language Model (LLM)? That it performs well on the relevant benchmarks---which hopefully measure, with some validity, the presence of capabilities that are also challenged in real application. But what makes the model perform well? What gives a model its abilities? We take a recently introduced type of benchmark that is meant to challenge capabilities in a goal-directed, agentive context through self-play of conversational games, and analyse how performance develops as a function of model characteristics like number of parameters, or type of training. 
We find that while there is a clear relationship between number of parameters and performance, there is still a wide spread of performance points within a given size bracket, which is to be accounted for by training parameters such as fine-tuning data quality and method. From a more practical angle, we also find a certain degree of unpredictability about performance across access methods, possible due to unexposed sampling parameters, and a, very welcome, performance stability against at least moderate weight quantisation during inference.
\end{abstract}

\section{Introduction}

Previous work 
has established that LLMs can be made to ``self-play'' dialogue games, and that their performance in doing so can be used as a differentiator between models \cite{chalamalasetti-etal-2023-clembench,qiao2023gameeval,li2023static,gong2023mindagent,wu2024smartplay,Zhou2024-sotopia,duan2024gtbench}. 
In particular,
\citet{chalamalasetti-etal-2023-clembench} have provided construct validity arguments for this approach that connect performance to the underlying constructs \textit{understanding} and \textit{reasoning}.
What this previous work has left unexplored is which model properties determine performance. This is what the present paper asks. Our theoretical interest is in the following: What drives performance? What makes a model a good model? (Sections~\ref{sec:first_observations}, \ref{sec:details}). Out of practical interest, and as a guide for researchers interested in tapping the agentive potential of LLMs, we also ask what influences the performance of a given model during inference. (Section~\ref{sec:host}).

\noindent
Before we turn to these questions, however, we briefly recap what zero- or few-shot dialogue game play requires from a model.

\begin{figure}[t]
  \centering
  \begin{prompt}
  { %
Let us play a game! I will give you a word, and your task is to describe the concept behind the word, but without using the word itself and also without using some other related words that I give you.\\
I will then give your description to your partner. Your partner will guess what the word was. I will then give you the guess that your partner made, and if it was wrong, you can give another clue, again taking care not to use the forbidden words.\\
You win if your partner correctly guesses the word.\\
Please start your description with DESCRIPTION:, and produce only one paragraph of text.\\
The word to describe is \{\{WORD\}\}, and the other forbidden words are \{\{TABOO\}\}.
}
\end{prompt}
\vspace*{-.5cm}
\caption{Zero-shot prompt template for inducing an agent that plays the `Taboo' game; modified from \citet{chalamalasetti-etal-2023-clembench}}
\vspace*{-.2cm}
\label{fig:taboo}
\end{figure}

\paragraph*{LLMs as Zero-Shot Agents}

Figure~\ref{fig:taboo} gives an example of how in this approach a game playing agent is induced from an LLM. We can distinguish several distinct elements in this prompt: First, the general goal of the game and the possible moves are explained (first three paragraphs); let us label this as $\mathcal{G}_g$. Second, specific instructions are given on how the response of the model is to be formulated, in order to count as a game move (last paragraph; $\mathcal{G}_f$). We call both together simply $\mathcal{G}$, and its instantiation with a specific goal (here, target and taboo words) $\mathcal{G}_x$.

Let us formalise the underlying decision problem of which move to take at any point in a Dialogue Game as a Markov Decision Process $(\mathcal{S}, \mathcal{A}, P, R)$, where $\mathcal{S}$ denotes the state of the game, $\mathcal{A}$ the action space, $P$ the transition function $\mathcal{S} \times \mathcal{A} \rightarrow \Delta(\mathcal{S})$, and $R$ the reward function $\mathcal{S} \times \mathcal{A} \rightarrow \mathbb{R}$.
What we are expecting from the model can then be described as follows: (implicitly) derive $P$ and $R$ from $\mathcal{G}_g$, $\mathcal{A} \in V^{max}$ from $\mathcal{G}_f$ (where $V$ is the model's vocabulary), a policy $\pi$ ($\mathcal{S} \rightarrow \mathcal{A}$) from $\mathcal{G}_g$ (such that following it optimizes the expected reward), and a \textit{state interpretation function} $S$ that takes the conversation so far ($\mathcal{G}_x, \mathcal{C}$) to a state $s \in \mathcal{S}$. A tall order indeed.

It is in the state interpretation function $S$ and the way it is induced by $\mathcal{G}_g$ (although this is our terminology) that \citet{chalamalasetti-etal-2023-clembench} locate the \textit{understanding} abilities of a model (for which they give further analyses into \textit{situation model}, \textit{discourse model}, etc.; see there), and \textit{format instruction following} abilities in the function that applies $\mathcal{G}_f$ to restrict the output space. The performance along these axes is measured by a \textit{quality score} and a \textit{percentage-played score}, respectively, and it is with this fine-grained instrument with which we will relate model characteristics and performance.

\section{Related Work}

Two lines of work are relevant for our questions. Work on \textit{scaling laws} for neural networks has shown that there is a power law relationship between the size of a neural network model and its performance \cite{DBLP:journals/corr/abs-2001-08361}. While the exact nature of the relationship in this (empirical) ``law'' is contested \cite{DBLP:journals/corr/abs-2203-15556}, the general observation has held up well \cite{DBLP:journals/corr/abs-2403-08540}. Recent observations point at a larger than previously accounted for the role of the \textit{quality} of the training data \cite{DBLP:journals/corr/abs-2302-13971, DBLP:journals/corr/abs-2307-09288, DBLP:journals/corr/abs-2404-14219}.

With the exception of the most recent work cited above, the work on scaling laws has mostly looked at simple performance measures like the loss on next-token prediction. The second line of work relevant here has, in contrast, been concerned with the dynamics of performance development across training and model parameters. \citet{DBLP:journals/corr/abs-2206-07682} described patterns that they observed of performance on certain tasks plotted against model size  as \textit{emergence} of abilities; an interpretation later supported by \citet{DBLP:journals/corr/abs-2206-04615}, and  more recently questioned by \citet{DBLP:journals/corr/abs-2309-01809}

Where this work has looked at performance metrics beyond prediction loss or perplexity, it has mostly used classic static NLP tasks. Our interest in the following is to use the recently introduced \textit{game-based evaluation} paradigm, where performance is measured via interactive tasks (in self-play), to investigate empirical relations and development dynamics with respect to parameters such as model size or type of training data and this type of performance.

\begin{figure}
    \centering
    \includegraphics[width=\linewidth, trim=520 0 0 0, clip]{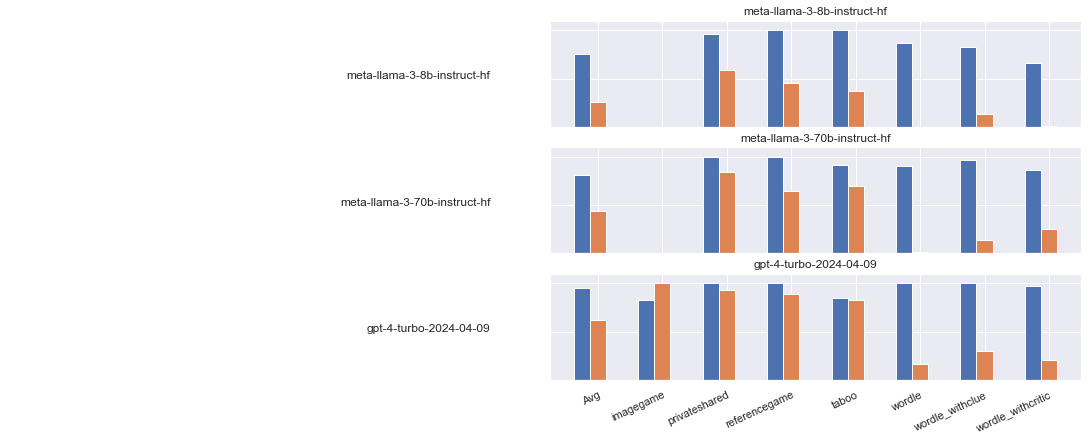}
    \vspace*{-.8cm}
    \caption{Profiles for three models. For each game (and the average), \%-played blue/left, quality brown/right. Top white line is at 100.}
    \label{fig:profiles}
    \vspace*{-.3cm}
\end{figure}

\section{The Starting Point: The Score Matrix}

As our measurement instrument, we use the \texttt{clembench} of \citet{chalamalasetti-etal-2023-clembench}, on account of it giving full access to the code, offering access to fine-grained measurements and the distinction between formal instruction following and competence (described above), and it being backed up by validity arguments.\footnote{Dialogue games contained in \texttt{clembench} are \texttt{taboo}, \texttt{imagegame} (\texttt{drawing}), \texttt{referencegame}, \texttt{wordle}, \texttt{wordle with clue} (\texttt{wordle+cl}), \texttt{wordle with critic} (\texttt{wordle+cr}), \texttt{private-shared} (\texttt{priv/sh}).}
An analysis with the same general structure could, of course, be performed using any score-giving benchmark.

We treat \texttt{clembench} as a function from model access to a score vector: $\mathcal{C}_\mathcal{G_X}: M \rightarrow \mathbf{s}$ (with each component of $\mathbf{s} \in [0, 1]$), where $\mathcal{C}$ is the \texttt{clemgame} framework, $\mathcal{G}_X$ is the set of game definitions and respective game instances, and $M$ is a model interface, in itself consisting of $d$, the inference/decoding interface, and $\Theta$, the set of weights. For some models, $d$ and $\Theta$ cannot vary independently, for access reasons (e.g., there is only one way of addressing GPT-4); for others, they can (e.g., accessing $\Theta_{\mathrm{llama-3-70B-instruct}}$ via an API or a local inference engine). We also test quantised versions of models so that a full specification can, e.g., \ be $(\mathrm{hf}, q_{4b}(\Theta_{l3-70B-inst}))$, for access to a 4bit quantised version of llama3-70b \cite{DBLP:journals/corr/abs-2307-09288} via Huggingface Transformers \cite{wolf-etal-2020-transformers}. 
The score vector $\mathbf{s}$ consists of two scores per game (measuring, as described above, format instruction following and game play ability), separate averages of these over all games, and a combined score that is the product, or the quality weighted by the percentage of attempted game plays not aborted through formally invalid moves; Figure~\ref{fig:profiles} visualises this for two models.
We took \texttt{clembench} version 1.6 \cite{beyer2024clembench2024} and ran the full benchmark on our cluster (4 x NVIDIA A100). We pair this score matrix with information about model characteristics (such as a number of parameters, size of the training corpus, and training methods), to the extent that we could reconstruct it.
See Appendix for the complete list of models, quantisation, and inference methods.

\section{First Observations} 
\label{sec:first_observations}

\begin{figure}
    \centering
    \hspace*{-.6cm}
    \includegraphics[width=1\linewidth]{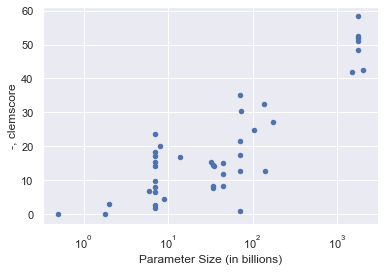}
    \vspace*{-.2cm}
    \caption{Plotting score vs.\ model size}
    \label{fig:size-v-score}
    \vspace*{-.3cm}
\end{figure}

Let us get the obvious out of the way. Figure~\ref{fig:size-v-score} plots (aggregated) score on \texttt{clembench} versus size of the model (in terms of number of parameters, log transformed; again with estimations where not public knowledge). The figure suggests that there might be a direct relationship, and indeed, a linear(-log) regression model finds a significant regression equation ($F(1, 18) = 23.56, p < 0.000$), with an $R^2$ of .57). If size of the training data is added in as a term, $R^2$ increases to .67 ($F(2,17) = 17.31, p < 0.000$). 

These results also indicate, however, that there is an unexplained remainder. This is brought out more clearly by Figure~\ref{fig:size-buckets-all} (Appendix), which visualises the spread within various model size classes. The knowledge that many of the models within one class often are derived from the same base model gives a good starting point for further analyses (see next section). 

As can be seen in Figures~\ref{fig:all-profiles} and \ref{fig:dynamics} (Appendix), while when aggregated it might look like \textit{\% played} (measuring format instruction following) and \textit{quality} (measuring the quality of game play in successfully terminated games) develop in lockstep, when looking at individual games, one can see that this need not be the case. The ability to play, both formally and with any success, the \textit{imagegame} (where player A instructs player B on how to draw a character-based grid-image) comes late (i.e., needs a larger model). In contrast, for \textit{taboo} or \textit{wordle}, the formatting instructions can be met even by smaller models, while quality increases only slowly.

Starting off of this, we look into these developments in more depth in the next section.

\section{From Base Model To Chat Model}
\label{sec:details}

\begin{table}[!htbp]
\centering
{\footnotesize
\begin{tabular}{TTl}
\hline
Base  Model                       & Derived Model                              & \multicolumn{1}{l}{Score} \\ \hline
                                  & starling-lm-7b-beta                         & 06.56                      \\
                                  & mistral-7b-instruct-v0.1                    & 08.01                      \\
                                  & openchat-3.5-0106                           & 17.10                      \\
                                  & openchat-3.5-1210                           & 18.22                     \\
\multirow{-5}{*}{\begin{tabular}[c]{@{}l@{}}mistral-7b\\ -v0.1\end{tabular}} & openchat-3.5                                & 23.64                     \\ \hline
                                  & llama-2-70b-chat-hf & 00.81                      \\
                                  & tulu-2-dpo-70b                              & 12.62                     \\
                                  & wizardlm-70b-v1.0                           & 17.40                      \\
\multirow{-4}{*}{\begin{tabular}[c]{@{}l@{}}llama-2\\ -70b\end{tabular}}     & sheep-duck-llama-2-70b-v1.1                 & 21.50                      \\ \hline
\end{tabular}%
}
\caption{Clemscore of different instruction-tuned models with the same base model}
\label{tab:same_base_main}
\end{table}

The fact that several of the models that were tested build off the same base model allows for a kind of `natural experiment' on the influence of tuning methods. 
As Figure~\ref{tab:same_base_main} shows, the scores achieved by the five derivatives from \texttt{mistral-7b-v0.1} are spread over 17 points. Where can this come from? We know what differs between these models (apart from the official \texttt{mistral-7b-instruct-v0.1} model, for which details are not available): \texttt{starling-lm-7b-beta} is trained using PPO  \cite{DBLP:journals/corr/abs-1909-08593}, while the \texttt{openchat-3.5} family is fine-tuned using C-RLFT  \cite{DBLP:journals/corr/abs-2309-11235}.
Interestingly, with respect to performance on this benchmark, the \texttt{openchat-3.5} models \textit{regressed} over time, with the newest version achieving 6.5 less than the oldest one. This might be a side-effect of the attempts by the developers to make \textit{coding + general tasks} capabilities and \textit{mathematical reasoning} capabilities separately accessible through separate chat templates \cite{DBLP:journals/corr/abs-2309-11235}.

We can also look at derivatives of \texttt{llama-2-70b-hf}, which similarly show a wide spread of performance points. Here, we focus on the data mix.
\model{Sheep-duck-llama-2-70b-v1.1}
\cite{platypus2023} 
combines Orca \cite{DBLP:journals/corr/abs-2306-02707} and Alpaca-inspired data,\footnote{\url{https://huggingface.co/Riiid/sheep-duck-llama-2-70b-v1.1}, \url{https://crfm.stanford.edu/2023/03/13/alpaca.html}}
\model{wizardlm-70b-v1.0} uses the Wizard Evol Instruct dataset \cite{DBLP:journals/corr/abs-2304-12244}, 
and \model{tulu-2-dpo-70b} employs the Tulu-v2 mix \cite{DBLP:journals/corr/abs-2311-10702}. Also, the latter model is trained with Direct Preference Optimization (DPO) \cite{DBLP:journals/corr/abs-2310-01377} using the ``UltraFeedback'' dataset \cite{DBLP:journals/corr/abs-2305-18290}. 

These observations further highlight the importance of fine-tuning data and method \cite{DBLP:journals/corr/abs-2404-14219}, showing that even small models can achieve comparatively high scores (e.g., \model{openchat-3.5}), provided that the right combination is found. It also highlights that generality might be harder to achieve, especially for smaller models: the \model{openchat-3.5} variants that perform worse here do indeed measure higher on other benchmarks \cite{DBLP:journals/corr/abs-2309-11235}.

\section{Practicalities: How to Host an Agent}
\label{sec:host}

Above, we have taken care to include in a full specification of a scored model (e.g., $(\mathrm{hf}, q_{4b}(\Theta_{l3-70B-inst}))$) both details about possible weight quantisation and about the access method. One might wonder why the latter---is a model not fully defined by its weights (and the network architecture)? As Table~\ref{tab:inference_diff} shows, perhaps surprisingly, the way of accessing a model (via API provider, locally) influences the achieved scores of around 5 clemscore points in the case of \model{meta-llama-3-70b-instruct}. For more discussion, see the Appendix; here, we point to one possible reason for the discrepancy: not all sampling parameters are exposed by these APIs, and hence, the actually used settings might differ.

\begin{table}[h]
\centering
{\footnotesize
\begin{tabular}{Tll}
\hline
Model & Backend & Score \\
\hline
Meta-Llama-3 & Groq & 39.34  \\
-70B-Instruct & Together AI & 35.20 \\ 
 & HuggingFace (local) & 35.11 \\ 
 & Anyscale & 34.26 \\ 
 \hline
Meta-Llama-3 & Together AI & 21.66 \\
-8B-Instruct & HuggingFace (local) & 19.99 \\ 
 & Anyscale & 19.32 \\ 
 & Groq & 17.79 \\ 
 \hline
\end{tabular}
}
\caption{Clemscore using different inference methods.}
\label{tab:inference_diff}
\end{table}

\begin{table}[h]
\centering
{\footnotesize
\begin{tabular}{lll}
\hline
Model  & Quantization & Score \\
\hline
Meta-Llama-3 & None & 19.99 \\
-8b-instruct & GGUF Q8\_0   & 20.54 \\
 & GGUF Q4\_K\_M & 11.75 \\
\hline
c4ai-command  & None & 24.94 \\
-r-plus & GGUF Q8\_0   & 25.53 \\
 & GGUF Q4\_K\_M & 19.48 \\
\hline
Meta-Llama-3  & None & 35.11 \\
-70b-instruct & GGUF Q8\_0 & 38.88 \\
& GGUF Q4\_K\_M & 33.57 \\
\hline
\end{tabular}
}
\caption{Clemscores at different quantization.}
\label{tab:quant_small}
\end{table}

Lastly, we found that model quantization also influences the \textit{clemscore}, as shown in Table \ref{tab:quant_small}, but not dramatically so, with 8-bit quantization having only 
negligible impact for the examined models.\footnote{Lower scores for the unquantized versions are likely due to the mentioned differences in sampling between HuggingFace \texttt{transformers} used for unquantized versions and \texttt{llama.cpp} used for quantized GGUF versions.} Smaller models, like Meta-Llama-3-8b-Instruct, show a greater impact on clembench performance at 4-bit quantization than larger models. Overall, we consider this good news, as it seems safe to serve 8bit quantized models (at lower cost) for agentive tasks.

\section{Conclusions}

We use a recently introduced benchmark, \model{clembench}, that test situated agency capabilities through LLM self-play and relates model performance to model characteristics. 
We find that while model scale, both in terms of model parameter size and training data amount, accounts for a large portion of performance on the benchmark, there is a notable impact of other factors. Comparing models trained off the same base in a kind of ``natural experiment'', we find that the quality of the instruction-tuning data mix also appears to profoundly impact performance, as models of the same magnitude show a wide spread of performance scores on the benchmark.
In particular, we find a good mix between ``chat'' and ``reasoning'' data in the fine-tuning data to be beneficial.
From a more practical perspective, we also find that the inference implementation has a notable influence, while conversely milder weight quantisation has little impact; this suggests best practices in setting up models for realising situated interactions.

\clearpage
\section*{Limitations}

All clembench game instances used are English only. While some of the models mentioned are able to process and generate text in other languages, our findings do not cover potential performance differences with non-English inputs.

For proprietary models accessed via remote API, information about training data and inference methods is not fully available. This limits our ability to compare them and derive conclusions from the comparison properly. See Appendix \ref{sec:app-inference} for details.

While clembench closely approximates natural text conversations, the produced dialogues are still artificial, as are the parsing and scoring of model outputs. Thus, high clembench scores are not to be seen as indicating good end-user performance or naturalness of conversation of the examined models.

The clembench score is also not an indication of the alignment or safety of a model.
Note that sometimes performance on clembench games might be at odds with safety mechanisms built into some of the models: A refusal to follow game instructions ("As an AI model, I can't...") would simply lead to a parsing failure, and a 0 'played' score for instance. However, as none of the game instructions could correctly be classified as unsafe requests, we do not consider this to be an unfair disadvantage for these models.

\bibliography{light_bulb-references,anthology}

\appendix
\footnotetext[1]{Our investigation on the observed behaviors is dependent on details marked by the model-provider/developers.}

\section{The Full Results Overview}

\begin{figure*}
    \centering
    \includegraphics[width=.48\linewidth, trim=150 0 0 0, clip]{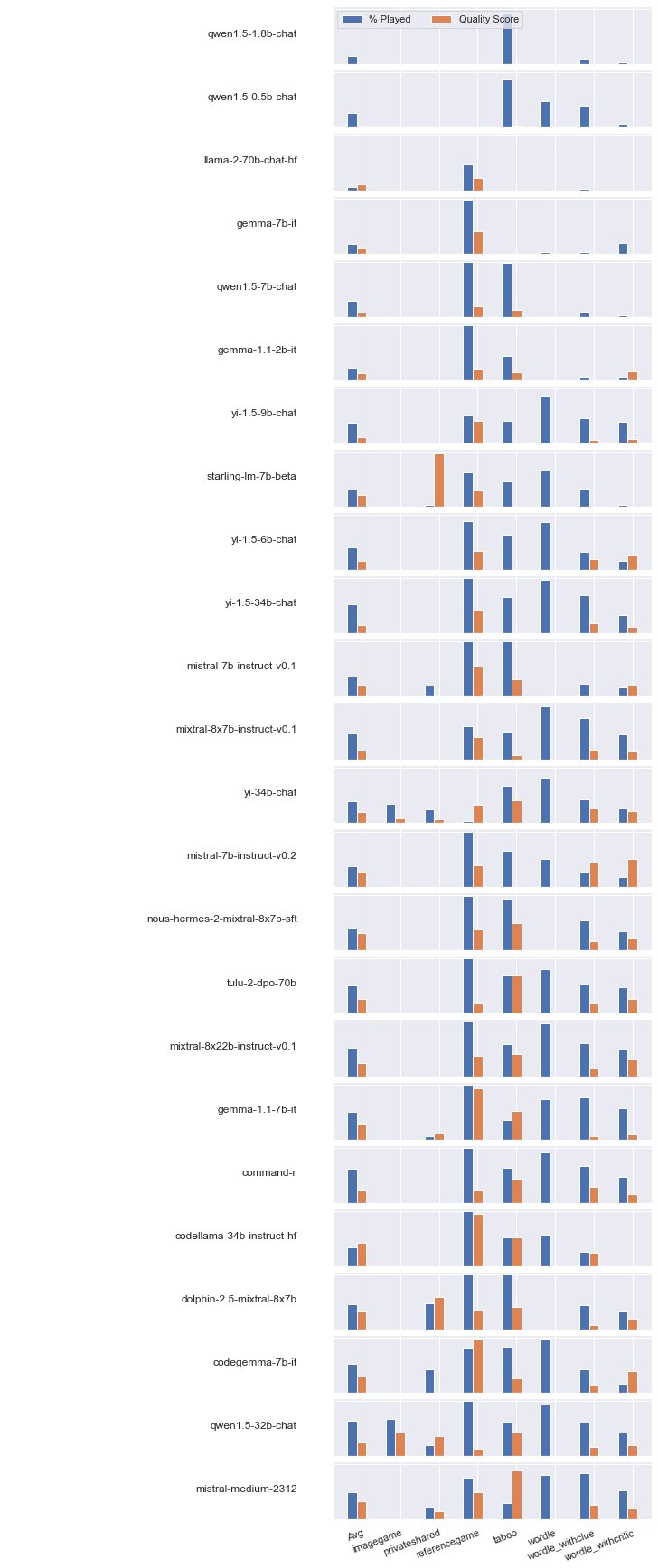}
    \includegraphics[width=.48\linewidth, trim=150 0 0 0, clip]{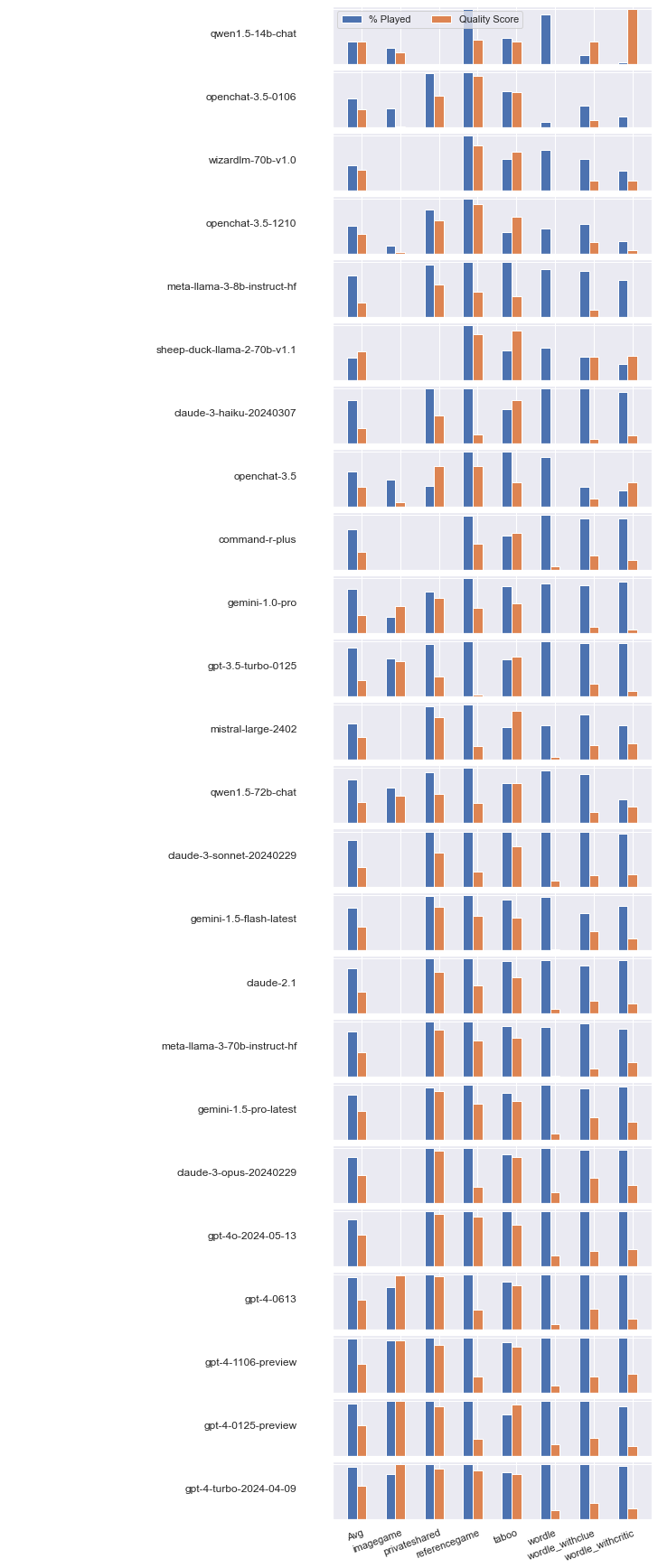}
    \caption{Profiles for all models (default inference method, not quantised), sorted by aggregated score (clemscore). For each game (and the average), \%-played blue/left, quality brown/right.}
    \label{fig:all-profiles}
\end{figure*}

\begin{figure*}
    \centering
    \hspace*{-.7cm}
    \includegraphics[width=1.1\linewidth]{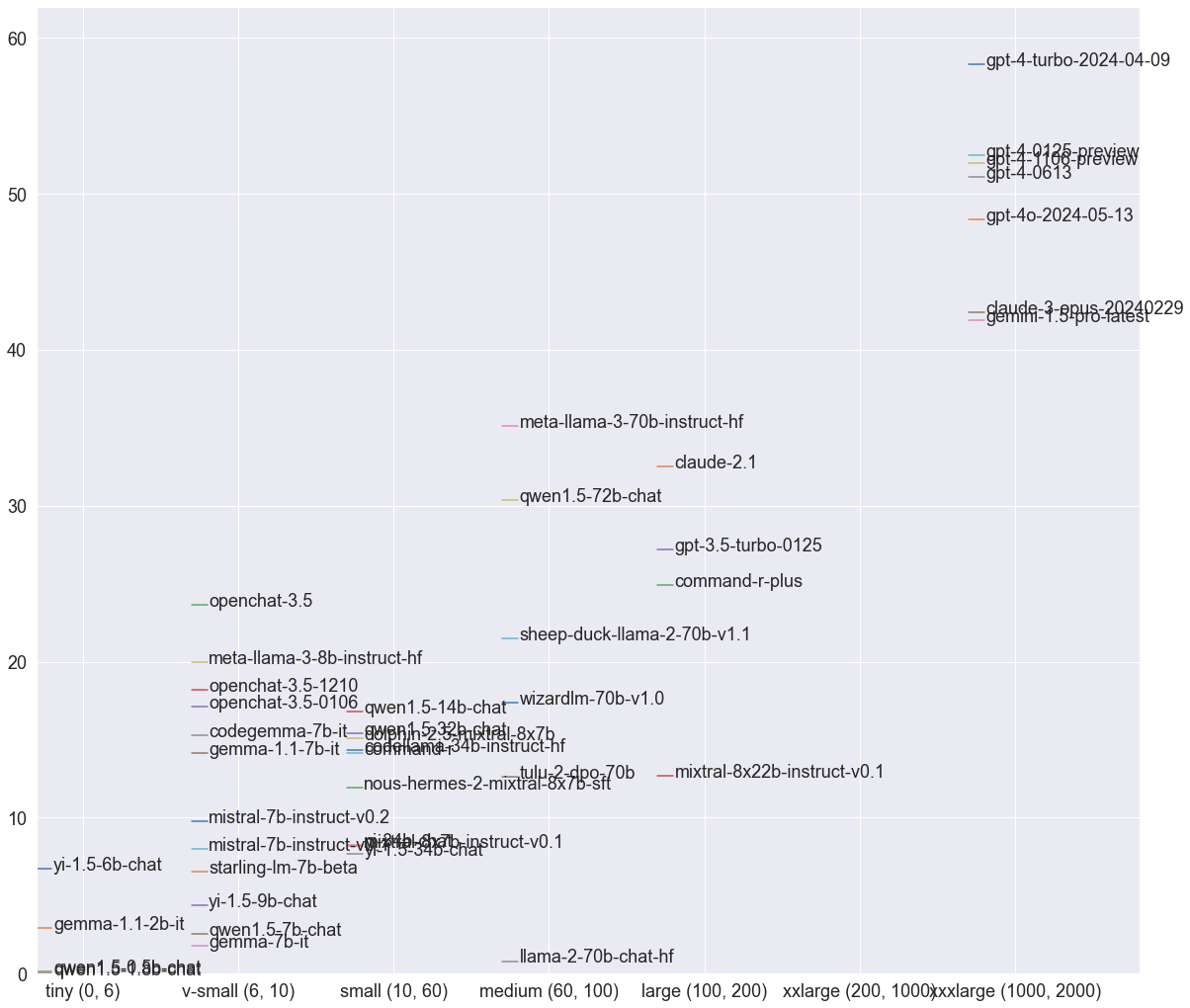}
    \caption{Performance, binned models sizes}
    \label{fig:size-buckets-all}
\end{figure*}

Figure~\ref{fig:all-profiles} shows aggregated scores and a breakdown by game for each model.  Figure~\ref{fig:size-buckets-all} shows the \textit{clemscore} of each model, where the models are binned in size buckets. This illustrates the spread of achievable performance within the same size bracket.

\section{Dynamics}

\begin{figure}
    \centering
    \hspace*{-.4cm}
    \includegraphics[width=1.1\linewidth]{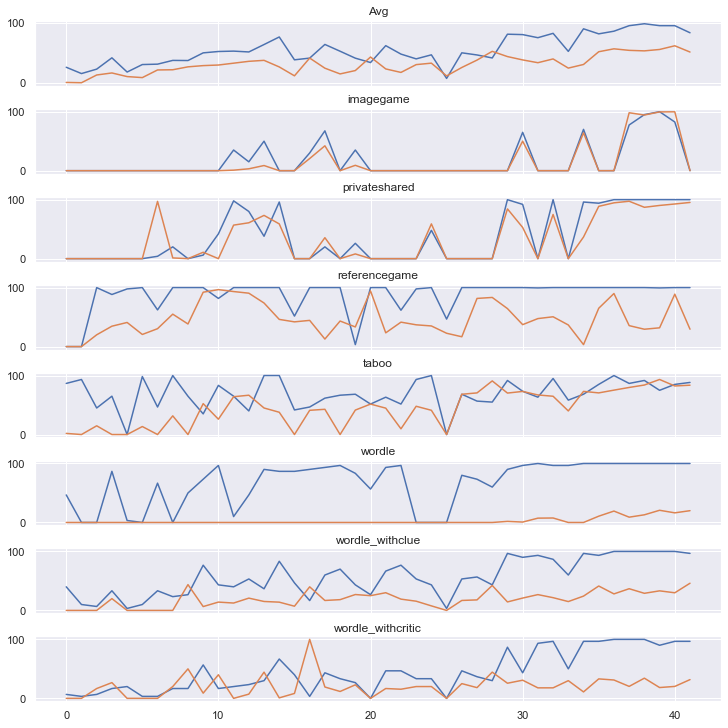}
    \vspace*{-.5cm}
    \caption{Dynamics of improvements. Models sorted by size (parameters) and, within same size, by aggregated score. I.e., on the left is the smallest and worst performing model, on the right the best performing of the biggest ones. Blue is \% played, orange is quality score.}
    \label{fig:dynamics}
\end{figure}

Figure~\ref{fig:dynamics} plots \textit{\% played} and \textit{quality} per model, with the models sorted by size, and within the same size, by performance. That is, by going from left to right on this graph, the improvement due to size (or other factors) can be seen. While for the aggregated measure in the top row, both metrics show a not very steep, but steady incline, looking at the individual games reveals difference between the games. The \textit{imagegame} sticks out as apparently requiring special capabilities (namely, being able to interpret character-sequences as ``images'') that are likely due to special training data and not model size. \textit{Taboo}, on the other hand, shows a nice steady improvement. Lastly, \textit{wordle} only begins to see modest improvements for very large models; the line suggests that here an increase of size might yield further improvements. None of these curves seem to point a sudden emergence of a capability at a certain size point.

\section{From Base Model to Chat Model}
\label{sec:app_derived}

\begin{table*}[h]
    \centering
    \hspace*{-1cm}
    {\scriptsize
    \setlength{\tabcolsep}{3pt}
    \begin{tabular}{TTllrrrp{.3\linewidth}}
\toprule
       Base Model &                     Model Name &        Model Provider & Release Date &    cs &  \% pl &  q sc &                                                                           Instruction Tuning Data \\
\midrule
         gemma-7b &                    gemma-7b-it &                Google &   2024-02-01 &  1.82 & 17.78 & 10.23 &                                                                                               NaN \\
         gemma-7b &                gemma-1.1-7b-it &                Google &   2024-04-01 & 14.14 & 49.67 & 28.46 &                                                                                               NaN \\
         gemma-7b &                codegemma-7b-it &                Google &   2024-04-01 & 15.30 & 51.95 & 29.45 &                                                                                               NaN \\
   llama-2-70b-hf &            llama-2-70b-chat-hf &                  Meta &   2023-07-01 &  0.81 &  7.14 & 11.31 &                                                                                               NaN \\
   llama-2-70b-hf &                 tulu-2-dpo-70b &               Allenai &   2024-11-01 & 12.62 & 49.76 & 25.37 &                                                      ultrafeedback-binarized, tulu-v2-sft-mixture \\
   llama-2-70b-hf &              wizardlm-70b-v1.0 &          Wizardlmteam &   2023-08-01 & 17.40 & 46.19 & 37.66 &                                                                    wizardlm-evol-instruct-v2-196k \\
   llama-2-70b-hf &    sheep-duck-llama-2-70b-v1.1 &                  Riid &   2023-09-01 & 21.50 & 41.19 & 52.20 &                                                                     orca and alpaca inspired data \\
  mistral-7b-v0.1 &            starling-lm-7b-beta &             Nexusflow &   2024-03-01 &  6.56 & 30.89 & 21.25 &                                                                                            nectar \\
  mistral-7b-v0.1 &       mistral-7b-instruct-v0.1 &             Mistralai &   2023-09-01 &  8.01 & 37.14 & 21.58 &                                                                                               NaN \\
  mistral-7b-v0.1 &              openchat-3.5-0106 &              Openchat &   2024-01-01 & 17.10 & 52.57 & 32.52 & sharegpt, openorca, capybara, goat, glaive, metamathqa, mathinstruct, oasst1, feedback-collection \\
  mistral-7b-v0.1 &              openchat-3.5-1210 &              Openchat &   2023-12-01 & 18.22 & 51.19 & 35.60 & sharegpt, openorca, capybara, goat, glaive, metamathqa, mathinstruct, oasst1, feedback-collection \\
  mistral-7b-v0.1 &                   openchat-3.5 &              Openchat &   2023-10-01 & 23.64 & 63.52 & 37.22 &                      sharegpt, openorca, capybara, goat, glaive, metamathqa, mathinstruct, oasst1 \\
mixtral-8x7b-v0.1 &     mixtral-8x7b-instruct-v0.1 &             Mistralai &   2023-12-01 &  8.17 & 47.62 & 17.15 &                                                                                               NaN \\
mixtral-8x7b-v0.1 & nous-hermes-2-mixtral-8x7b-sft &          NousResearch &   2024-01-01 & 11.95 & 39.68 & 30.12 &                                                                                    openhermes-2.5 \\
mixtral-8x7b-v0.1 &       dolphin-2.5-mixtral-8x7b & cognitivecomputations &   2023-12-01 & 15.10 & 46.38 & 32.55 &                              magicoder-oss-instruct-75k, magicoder-evol-instruct-110k, openhermes \\
\bottomrule
\end{tabular}

    }
    \caption{Comparing models trained off the same base model}
    \label{tab:same_base}
\end{table*}

\begin{table*}[h]
    \centering
    \hspace*{-1cm}
    {\scriptsize
    \setlength{\tabcolsep}{3pt}
    \begin{tabular}{ll}
\toprule
Dataset Name & Links \\
\midrule
capybara & \url{https://huggingface.co/datasets/LDJnr/Capybara} \\
feedback-collection & \url{https://huggingface.co/datasets/prometheus-eval/Feedback-Collection} \\
glaive & \url{https://huggingface.co/datasets/glaiveai/glaive-code-assistant} \\
goat & \url{https://huggingface.co/datasets/tiedong/goat} \\
magicoder-evol-instruct-110k & \url{https://huggingface.co/datasets/ise-uiuc/Magicoder-Evol-Instruct-110K} \\
magicoder-oss-instruct-75k & \url{https://huggingface.co/datasets/ise-uiuc/Magicoder-OSS-Instruct-75K} \\
mathinstruct & \url{https://huggingface.co/datasets/TIGER-Lab/MathInstruct} \\
metamathqa & \url{https://huggingface.co/datasets/meta-math/MetaMathQA} \\
nectar & \url{https://huggingface.co/datasets/berkeley-nest/Nectar} \\
oasst1 & \url{https://huggingface.co/datasets/OpenAssistant/oasst_top1_2023-08-25} \\
openhermes & \url{https://huggingface.co/datasets/teknium/openhermes} \\
openhermes-2.5 & \url{https://huggingface.co/datasets/teknium/OpenHermes-2.5} \\
openorca & \url{https://huggingface.co/datasets/Open-Orca/OpenOrca} \\
sharegpt & \url{https://huggingface.co/datasets/openchat/openchat_sharegpt4_dataset} \\
tulu-v2-sft-mixture & \url{https://huggingface.co/datasets/allenai/tulu-v2-sft-mixture} \\
ultrafeedback-binarized & \url{https://huggingface.co/datasets/HuggingFaceH4/ultrafeedback_binarized} \\
wizardlm-evol-instruct-v2-196k & \url{https://huggingface.co/datasets/WizardLMTeam/WizardLM_evol_instruct_V2_196k} \\
\bottomrule
\end{tabular}

    }
    \caption{List of instruction tuning datasets}
    \label{tab:same_base}
\end{table*}

Generally, when a model is improvised by its developers, it is majorly aimed to allow the model to have improved capabilities with regards to overall improvements on the downstream benchmarks. 
However, our analysis says the opposite. As we outline this in Figure \ref{fig:size-v-score}, \texttt{openchat-3.5} (Jun 2023) has a better clemscore compared to its successor counterparts i.e. \texttt{openchat-3.5-1210} (Dec 2023), and \texttt{openchat-3.5-0106} (Jan 2024). All three models are fine-tunes of the same base model, i.e. \texttt{mistral-7b-v0.1}.  

Is model improvement across benchmarks proportional to improvement on agentive tasks? If so, are there additional factors that can alter this relationship?\footnotemark[1]

Below, we analyze potential factors responsible for the observed deviations:

First, looking into the openchat series of models, they are instruction tuned using the C(Conditioned)-RLFT technique \cite{DBLP:journals/corr/abs-2309-11235}, which eases the need for high-quality preference optimization datasets---and uses SFT training data, consisting of a small amount of expert data without any preference labels. So, to put it aside, all of these models utilize a similar tuning technique, thus it should be fine to disregard the factor of tuning method from the discussion. 

Next, selection of training data, might also have an effect on the models' performance. Before that, we would like to emphasize on the formatting of input data for either of these models. Unlike \texttt{openchat-3.5}, which uses a single prompt format for all training inputs, later versions are trained to handle two different formats or "modes": one for general reasoning tasks and another for mathematical reasoning. All three models are mostly trained on similar data, which is mainly synthetic multi-turn conversations (see table 5). Apparently, the key difference is that the later models are also trained to evaluate and give feedback on responses.

Conclusively, even though this can be regarded as speculation, based on the observations indicated through the above discussion, it would be ideal to assert that an increase in model competencies inversely affects precision on tasks at hand (i.e., individual dialogue games). Smaller models can compress information to a certain extent, and overloading them with more capabilities may reduce proficiency.

More importantly, in clembench, for individual game instances, which seek an ensemble of abilities from the testing agent, it is inefficient to identify them explicitly beforehand. This comes as a drawback, since models like \texttt{openchat-3.5-1210}, and \texttt{openchat-3.5-0106} (available under the category of \texttt{v-small} models) are trained to process abilities like general reasoning and mathematical reasoning separately.

\section{Inference}
\label{sec:app-inference}

\begin{table*}[!htbp]
\resizebox{\textwidth}{!}{%
\begin{tabular}{llllll}
\hline
Model                                      & \begin{tabular}[c]{@{}l@{}}API Provider / \\ Backend\end{tabular} & Clemscore & Avg. \% Played & \begin{tabular}[c]{@{}l@{}}Avg. \% Quality \\ Score\end{tabular} & Link                                                                   \\ \hline
\multirow{4}{*}{\texttt{Meta-Llama-3-70B-Instruct}} & HuggingFace (local)     & 35.11     & 80.72             & 43.5                     & \url{https://huggingface.co/meta-llama/Meta-Llama-3-70B-Instruct}      \\
                                           & Groq                  & 39.34     & 82.35             & 47.77                    & \url{https://groq.com/}                                                \\
                                           & Anyscale              & 34.26     & 80.00                & 42.82                    & \url{http://anyscale.com/}                                        \\
                                           & Together AI           & 35.20      & 79.52             & 44.26                    & \url{https://www.together.ai/}                                    \\ \hline
\multirow{4}{*}{\texttt{Meta-Llama-3-8B-Instruct}}  & HuggingFace (local)     & 19.99     & 76.1              & 26.27                    & \url{https://huggingface.co/meta-llama/Meta-Llama-3-8B-Instruct} \\
                                           & Groq                  & 17.79     & 77.43             & 22.98                    & \url{https://groq.com/}                                           \\
                                           & Anyscale              & 19.32     & 75.81             & 25.48                    & \url{http://anyscale.com/}                                        \\
                                           & Together AI           & 21.66     & 74.67             & 29.01                    & \url{https://www.together.ai/}                                    \\ \hline
\end{tabular}%
}
\caption{Inference results of the used models (specific to API Provider/Backend) on the clembench. Reports for each model the clemscore, with average \% played, and average \% quality score. For models acquired through HuggingFace, the corresponding repository is linked, for those accessed via a (remote) inference interface, the API provider's website.}
\label{tab:backend_results}
\end{table*}

\begin{table*}[!htbp]
\resizebox{\textwidth}{!}{%
\begin{tabular}{llllll}
\hline
Model Name                        & File Format & Clemscore & Avg. \% Played & \begin{tabular}[c]{@{}l@{}}Avg. \% Quality \\ Score\end{tabular} & Link                                                                      \\ \hline
\texttt{meta-llama-3-8b-instruct-gguf-q4}  & GGUF Q4\_K\_M & 11.75     & 54.84          & 21.43                 & \url{https://huggingface.co/MaziyarPanahi/Meta-Llama-3-8B-Instruct-GGUF}  \\
\texttt{meta-llama-3-8b-instruct-gguf-q8}  & GGUF Q8\_0   & 20.54     & 69.05          & 29.74                 & \url{https://huggingface.co/MaziyarPanahi/Meta-Llama-3-8B-Instruct-GGUF}  \\
\texttt{meta-llama-3-8b-instruct-hf}       & safetensors & 19.99     & 76.1           & 26.27                 & \url{https://huggingface.co/meta-llama/Meta-Llama-3-8B-Instruct}          \\ \hline
\texttt{c4ai-command-r-plus-gguf-q4}       & GGUF Q4\_K\_M & 19.48     & 66.43          & 29.33                 & \url{https://huggingface.co/pmysl/c4ai-command-r-plus-GGUF}               \\
\texttt{c4ai-command-r-plus-gguf-q8}       & GGUF Q8\_0   & 25.53     & 74.24          & 34.39                 & \url{https://huggingface.co/pmysl/c4ai-command-r-plus-GGUF}               \\
\texttt{command-r-plus}                    & -           & 24.94     & 74.9           & 33.3                  & \url{https://cohere.com/}                                                 \\ \hline
\texttt{meta-llama-3-70b-instruct-gguf-q4} & GGUF Q4\_K\_M & 33.57     & 78.33          & 42.86                 & \url{https://huggingface.co/MaziyarPanahi/Meta-Llama-3-70B-Instruct-GGUF} \\
\texttt{meta-llama-3-70b-instruct-gguf-q8} & GGUF Q8\_0   & 38.88     & 74.4           & 52.26                 & \url{https://huggingface.co/MaziyarPanahi/Meta-Llama-3-70B-Instruct-GGUF} \\
\texttt{meta-llama-3-70b-instruct-hf}      & safetensors & 35.11     & 80.72          & 43.5                  & \url{https://huggingface.co/meta-llama/Meta-Llama-3-70B-Instruct}         \\ \hline
\end{tabular}%
}
\caption{Inference results of the used models (specifically quantized or with normal floating precision) on the clembench. Reports for each model clemscore, with average \% played, and average \% quality score. For models acquired through HuggingFace, the corresponding repository is linked, for those accessed via a (remote) inference interface, the API provider's website.}
\label{tab:quantizations_results}
\end{table*}

Table~\ref{tab:backend_results} shows the surprising spread of scores achieved by what should be the same model, when accessed via different providers.

Three main factors might lead to these differences:

\textbf{Input formatting}, as in the applied chat template \citep{Carrigan2023hfblogchattemplates} or additions to the initial user or system message \citep{chawdhury2024content, lyu2024keeping}, can strongly impact the produced replies. As 'OpenAI-compatible' remote APIs only receive an array of message objects, we could not control for the actual input contexts. For our local inference, the method of creating input contexts was kept constant.

\textbf{Sampling parameters}, with only a subset commonly exposed via remote API, can also influence replies \citep{shi2024thorough} and are commonly tweaked to align better with end-user expectations. While we controlled for temperature, setting it to zero for local inference and remote APIs, other common sampling parameters like repetition penalties were left to default values supplied by model creators or API providers.

Lastly, inference features pertaining to \textbf{turn-taking} can have great influence on clembench 'played' scores, such as the proper production and processing of end-of-turn tokens to stop further text generation\footnote{Clembench uses tokens predefined by model developers to stop text generation, and while these are metadata, we consider them to be part of model architecture and inference implementation. Generation stopping is important for user experience and saves resources, thus this sensitivity is desirable over the use if custom stopping measures by a benchmark.}. A prominent example of this are the recent Meta-Llama-3-Instruct models, the highly popular HuggingFace implementation of which had the end-of-text token, used mainly for non-instruct uses and training, set as the stop token instead of the end-of-turn token, leading to overly long replies and quick deterioration of the generated text \citep{hf2024llama3eot}.

The inaccessibility of many of these inference parameters on commercial remote APIs, along with proprietary secrecy about unexposed parameters and training, make it hard to argue about their performance effects beyond model scale as mentioned in Section \ref{sec:first_observations}. There is also a noticeable sparsity of literature examining the effects of these factors on benchmark scores, with API parameter inaccessibility likely exacerbated by prohibitively high local computational demands posed by the stochastic nature of sampling-based generation.

\ \\
Table~\ref{tab:quantizations_results} shows results at different quantisation strengths.

\end{document}